\documentclass{article}
\usepackage{amssymb}
\usepackage{ijcai25}

\usepackage{times}
\usepackage{wrapfig}
\usepackage{soul}
\usepackage{url}
\usepackage[hidelinks]{hyperref}
\usepackage[utf8]{inputenc}
\usepackage[small]{caption}
\usepackage{graphicx}
\usepackage{amsmath}
\usepackage{amsthm}
\usepackage{booktabs}
\usepackage{algorithm}
\usepackage{algorithmic}
\usepackage[switch]{lineno}
\usepackage{caption}
\usepackage{subcaption}
\usepackage{booktabs}
\usepackage{hyperref}
\usepackage{multirow}
\usepackage{soul}
\usepackage{enumitem}
\usepackage{amsmath,amsfonts}
\usepackage{algorithmic}
\usepackage{graphicx}
\usepackage{textcomp}
\usepackage{xcolor}
\usepackage{wrapfig}
\usepackage{threeparttable}
\usepackage{color}
\usepackage{algorithmic}
\usepackage{algorithm}
\usepackage{balance}
\usepackage{amssymb}
\usepackage{hyperref}
\usepackage{makecell}

\title{HCRide: Harmonizing Passenger Fairness and Driver Preference for Human-Centered Ride-Hailing}

\author{
Lin Jiang\textsuperscript{1},
Yu Yang\textsuperscript{2},
Guang Wang\textsuperscript{1}\thanks{Corresponding author}\\
\affiliations
\textsuperscript{1}Department of Computer Science, Florida State University\\
\textsuperscript{2}Department of Computer Science and Engineering, Lehigh University\\
\emails
lj23d@fsu.edu, yuyang@lehigh.edu, guang@cs.fsu.edu
}

\begin{document}

\maketitle

\begin{abstract}
Order dispatch systems play a vital role in ride-hailing services, which directly influence operator revenue, driver profit, and passenger experience. Most existing work focuses on improving system efficiency in terms of operator revenue, which may cause a bad experience for both passengers and drivers. Hence, in this work, we aim to design a human-centered ride-hailing system by considering both passenger fairness and driver preference without compromising the overall system efficiency. However, it is nontrivial to achieve this target due to the potential conflicts between passenger fairness and driver preference since optimizing one may sacrifice the other. To address this challenge, we design HCRide, a \textbf{H}uman-\textbf{C}entered \textbf{Ride}-hailing system based on a novel multi-agent reinforcement learning algorithm called \textbf{H}armonization-oriented \textbf{A}ctor-\textbf{Bi}-\textbf{C}ritic (Habic), which includes three major components (i.e., a multi-agent competition mechanism, a dynamic Actor network, and a Bi-Critic network) to optimize system efficiency and passenger fairness with driver preference consideration. We extensively evaluate our HCRide using two real-world ride-hailing datasets from Shenzhen and New York City. Experimental results show our HCRide effectively improves system efficiency by 2.02\%, fairness by 5.39\%, and driver preference by 10.21\% compared to state-of-the-art baselines.

\end{abstract}

\footnotetext[1]{Accepted to the IJCAI 2025 (Human-Centred AI Track).}

\section{Introduction}
In recent years, ride-hailing services (e.g., Uber, Lyft, Ola Cabs, and DiDi Chuxing) have become indispensable to our daily transportation needs. By 2023, the global ride-hailing market size was valued at \$109.3 billion, and it is expected to expand at
a growth rate of 12.70\% from 2024 to 2033 \cite{MarketScale}. One of the most important components of ride-hailing services is the order dispatch system, which directly impacts the revenue of platforms, the work experience of drivers, and the user experience of passengers.

\begin{figure*}[t]\centering
    \centering
    \includegraphics[width=0.95\linewidth, keepaspectratio=true]{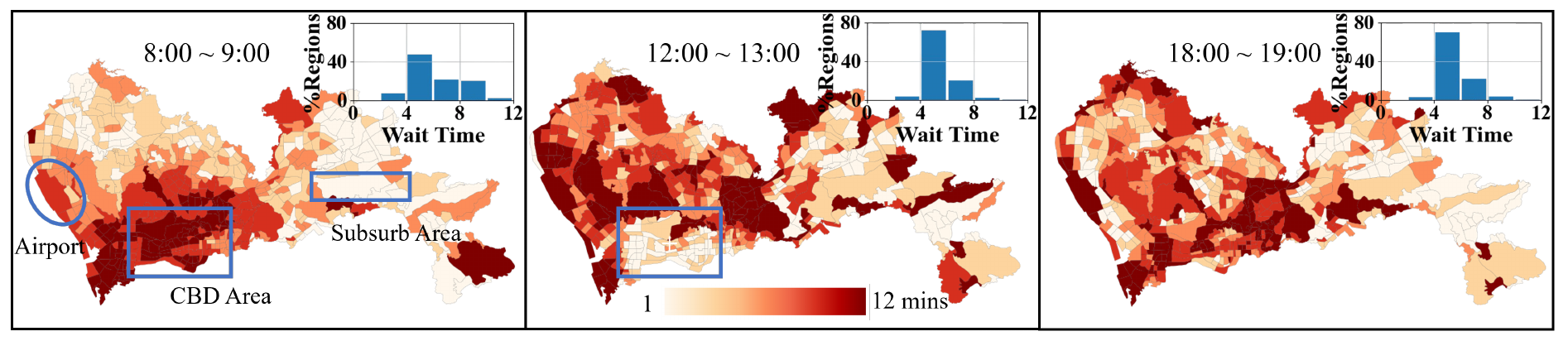}
    \caption{Average Passenger Waiting Time in Different Regions at Different Periods}
    \label{fig:spatial distribution}
    \vspace{-5pt}
\end{figure*}

Due to its importance, order dispatch has attracted significant attention from both industry and academia~\cite{chen2019dispatching,yuan2021real,xu2018large,wang2022fed}. However, most existing works focus on maximizing system revenue, which can potentially compromise the driver and passenger experience. Although some recent studies consider passenger fairness~\cite{suhr2019two,zhou2023optimal,nanda2020balancing,wang2021data,wang2023fairmove,jiang2023faircod}, most adopt an \emph{absolute fairness} setting~\cite{wang2022brief}, assuming all passengers should experience equal waiting times regardless of location, which overlooks the dynamic nature of supply and demand across regions. Furthermore, driver preferences are frequently ignored, leading to poor experiences when drivers are dispatched to unfamiliar or undesired areas.

Hence, in this work, we aim to design a human-centered ride-hailing order dispatch system that harmonizes passenger fairness and driver preference. However, it is nontrivial to achieve this due to the following two reasons. (i) It is challenging to formally define passenger fairness and driver preference since they have highly spatial and temporal dynamics. (ii) Harmonizing passenger fairness and driver preference is also challenging due to their potential conflicts since improving passenger fairness may not align with drivers' preferences. For example, to ensure passenger fairness, drivers might be dispatched to high-demand areas, which could be distant from their preferred locations or lead to extended working hours beyond their preferred schedules.

To address the above challenges, we propose a \textbf{H}uman-\textbf{C}entered \textbf{Ride}-hailing framework, called \textbf{HCRide}, which aims to minimize total passenger waiting time and enhance fairness without compromising driver preferences. In HCRide, we formulate the order dispatch problem as a Constrained Markov Decision Process (CMDP), where the passenger fairness-aware reward serves as the optimization objective and the accumulated driver preference-based cost is treated as a constraint. Passenger fairness is formally defined based on the divergence of waiting times across different spatial-temporal contexts, considering both inter-region and intra-region levels. Driver preferences are modeled using each driver's historical visitation frequency to various regions, reflecting their working habits and regional familiarity. To solve this CMDP, we develop a novel multi-agent reinforcement learning (RL) algorithm called \textbf{H}armonization-oriented \textbf{A}ctor-\textbf{Bi}-\textbf{C}ritic (Habic). Habic consists of three key components. First, a multi-agent competition mechanism transforms the large joint action space into smaller distributed action spaces among a limited number of candidate agents. This enables a micro-level decision process~\cite{qin2022reinforcement} that helps manage competition among proximate heterogeneous drivers with varying preferences. Second, a Bi-Critic module incorporates two evaluation networks: one estimates the reward value reflecting system efficiency and passenger fairness, while the other assesses the cost related to driver preferences. Third, an Actor module utilizes the outputs of the Bi-Critic networks to generate dispatch decisions that balance both passenger fairness and driver preferences.

The key contributions of this work are as follows:
\begin{enumerate}

    \item To our knowledge, this is the first study on human-centered ride-hailing order dispatch that considers both passenger fairness and driver preference. Our design is motivated by social studies and data-driven analysis, from which we observe: (i) a notable discrepancy in waiting times among passengers, both within and across regions; (ii) drivers show distinct preferences for operational regions—some favor smaller, confined areas, while others are willing to cover broader locations.

    \item Based on the data-driven findings, we design a human-centered ride-hailing order dispatch system called HCRide to improve passenger fairness without compromising driver preferences. Spatio-temporal-aware fairness and preference are defined. The core of HCRide is a novel multi-agent RL algorithm called Habic, which includes a multi-agent competition mechanism and an Actor-Bi-Critic module to harmonize passenger fairness and driver preference.
    
    \item More importantly, we implement and extensively evaluate our HCRide based on two real-world ride-hailing datasets. Experiment results show our HCRide effectively improves system efficiency by 1.77\% and 2.02\%, inter-region fairness by 5.29\% and 5.28\%, intra-region fairness by 7.65\% and 5.39\%, and driver preference by 7.77\% and 10.21\% compared to baselines on the Shenzhen and NYC datasets, respectively. To verify our work, we have the code available at \href{https://github.com/LinJiang18/HCRide} {GitHub} \footnote{https://github.com/LinJiang18/HCRide}.

\end{enumerate}

 \begin{figure}[htbp]\centering
     \centering
\includegraphics[width=0.85\linewidth, keepaspectratio=true]{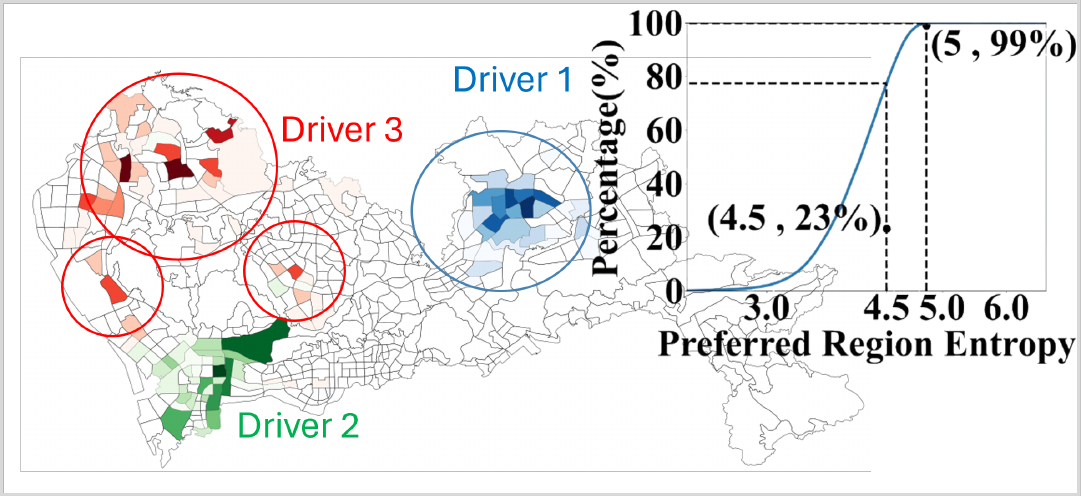}
     \caption{A Visualization of Driver Preference}
     \label{fig:driver preference}
 \end{figure}

\section{Socially Informed Fairness and Preference Formulation} \label{Data Analysis and Motivation}

\subsection{Data-driven Findings}
In our previous project, we conducted qualitative studies to understand people's perceptions of current ride-hailing services, focusing on their views on fairness and their personal preferences \cite{wang2025mixed}. In this paper, we also utilize real-world data with over one million ride-hailing orders to verify findings from a quantitative perspective, and finally, we have the following conclusions: 

1. There is strong demand among passengers for equitable waiting times, particularly relative to others in close spatial-temporal proximity. However, substantial disparities in waiting times exist both within and across regions, driven by various spatial-temporal factors. As shown in Fig.~\ref{fig:spatial distribution}, we visualize average passenger waiting times across 491 regions in Shenzhen during three periods: morning and evening rush hour, and noon non-rush hour. The results show that (i) \textbf{spatially}, significant differences exist between regions, such as between the Central Business District (CBD) and suburban areas; and (ii) \textbf{temporally}, even within the same region, average waiting times vary considerably within and across time periods.

2. Drivers have also reported having individual preferences for operating in specific areas at different times, such as locations near their homes, airports, or downtown districts. However, they are often assigned orders outside these preferred regions or in unfamiliar areas, which can negatively affect their satisfaction and operational efficiency. As shown in Fig.~\ref{fig:driver preference}, darker colors represent areas frequently visited by a given driver. We observe that some drivers (e.g., Driver 1) tend to operate within limited regions, while others (e.g., Driver 3) cover a broader range. The preferred region entropy analysis in the upper right of Fig.~\ref{fig:driver preference} further highlights the diversity in drivers’ operational areas.

\subsection{Design of Fairness and Preference} \label{Sec3-PROBLEM FORMULATION}

\subsubsection{\textbf{Passenger Fairness}} 

Motivated by the above findings, we define passenger fairness from both inter-region and intra-region perspectives, incorporating spatial and temporal patterns. In particular, intra-region fairness is defined as:
\begin{equation} \label{eq:2}
     WT(p_1 \mid u,v) = WT(p_2 \mid u,v)
\end{equation}
\noindent where \( WT(\cdot) \) denotes the passenger waiting time, and \( p_1 \) and \( p_2 \) are two passengers located in the same region \( u \) during a specific time period \( v \) (e.g., 8:00–9:00). In this study, each time period is set to one hour, implying that passengers within the same region and period should experience comparable waiting times. Depending on the application, the temporal granularity can be adjusted. Under Eq.~\ref{eq:2}, fairness is achieved when the waiting times for both passengers are equal.

For passengers in different regions, we define a fairness benchmark, denoted as \( WT_c(u,v) \), representing the expected average waiting time in region \( u \) during time period \( v \). A passenger’s waiting time in region \( u \) is considered fair if it aligns closely with this benchmark \( WT_c(u,v) \). Inspired by the concept of Demographic Parity~\cite{singh2018fairness}, we evaluate fairness across regions by comparing their respective fairness benchmarks as follows:
\begin{equation}
    \frac{WT_c(u_1,v)}{|C_1|} = \frac{WT_c(u_2,v)}{|C_2|}
\end{equation}
\noindent where \( |C_i| = \beta_i \times \frac{N^i_{\text{passenger}}}{N^i_{\text{driver}}} \); \( N_{\text{passenger}} \) is the historical average number of passengers in region \( u \) during period \( v \), \( N_{\text{driver}} \) denotes the number of drivers who prefer to operate in region \( u \), and \( \beta_i \) is an adjustable hyperparameter. Under this setting, the inter-region fairness benchmark \( WT_c(u,v) \) is inversely proportional to the supply-demand ratio. The underlying \textbf{rationale} is that the expected waiting time in a region should decrease when driver supply exceeds passenger demand.

\subsubsection{\textbf{Driver Preference}}
To quantify driver preferences, we introduce the following definitions:  
(i) \(\mathcal{U}\): the set of all regions in the city;  
(ii) \(\mathcal{H}^{+}_k\): the set of regions that provide positive feedback when driver \(k\) serves a passenger there;  
(iii) \(\mathcal{H}^{0}_k\): the set of regions that are neutral—i.e., do not yield positive feedback but are still acceptable to driver \(k\);  
(iv) \(\mathcal{H}^{-}_k\): the set of regions that result in negative feedback for driver \(k\). 
The formal representations of these sets are defined as follows:
\begin{align}
    & \mathcal{H}^{+}_k = \{u \in \mathcal{U} \, | \, V_k(u) > d\} \\
    & \mathcal{H}^{0}_k = \{u \in \mathcal{U} \, | \, dis(u,u_1) < \kappa V_k(u_1) \quad \forall u_1 \in \mathcal{H}^{+}_k \} \\
    & \mathcal{H}^{-}_k = \{u \in \mathcal{U} \, | \, u \notin \mathcal{H}^{+}_k \cup \mathcal{H}^{0}_k \}
\end{align}
\noindent where $V_k(u)$ denotes the historical visitation frequency of driver $k$ to region $u$. A region $u$ is classified into the Positive region set $\mathcal{H}^{+}_k$ for driver $k$ if the visitation frequency $V_k(u)$ exceeds a threshold $d$.
$dis(u,u_1)$ represents the distance between the region $u$ and $u_1$, which should be shorter than $\kappa V_k(u_1)$. Here, $\kappa V_k(u_1)$ defines the radius of influence for a positive region $u_1$. This means that if region $u$ falls within the influence radius of the positive region $u_1$, it is deemed acceptable for driver $k$ to operate in.
 
In Fig.~\ref{fig:driver preference}, we illustrate examples of the influence radius for drivers 1 and 3. The negative region set $\mathcal{H}^{-}_k$ comprises all other regions that are not included in the positive region set $\mathcal{H}^{+}_k$ or the neutral region set $\mathcal{H}^{0}_k$.

\section{HCRide System Design} \label{Sec5-PROPOSED FRAMEWORK}

In this part, we introduce the detailed design of the human-centered ride-hailing order dispatch system HCRide, which prioritizes passenger fairness and accommodates driver preferences, considering various spatio-temporal factors.

\subsection{Order Dispatch Problem Formulation} \label{Sec:3.3}
Formally, we model the fairness-oriented, preference-aware ride-hailing order dispatch problem as a Constrained Markov Decision Process (CMDP) \(\mathcal{G}\), defined as an 8-tuple: $\mathcal{G} = \{\mathcal{S}, \mathcal{A}, \mathcal{P}, \mathcal{R}, \mathcal{C}, \mu, \gamma_r, \gamma_c\}$, where \(\mathcal{S}\) is the state space; \(\mathcal{A}\) the action space; \(\mathcal{P}\) the transition probability function, \(\mathcal{P}: \mathcal{S} \times \mathcal{A} \times \mathcal{S} \rightarrow [0,1]\); \(\mathcal{R}\) the reward function; \(\mathcal{C}\) the set of cost functions; \(\mu: \mathcal{S} \rightarrow [0,1]\) the initial state distribution; and \(\gamma_r, \gamma_c \in (0,1]\) are discount factors for future rewards and costs. Let \(\Pi\) denote the set of all stationary policies, and let \(\pi_\theta(a \mid s) \in \Pi\) be a policy parameterized by \(\theta\) that maps state \(s\) to a distribution over actions \(a\). We divide each day into consecutive time slots (e.g., one minute per slot) and perform state transitions from time slot \(t\) to \(t+1\). Dispatch decisions are executed at the beginning of each slot. The detailed formulation of CMDP \(\mathcal{G}\) is presented below.

\begin{itemize} 
    \item \textbf{Agent}: In our problem, we define each driver as an agent. An agent is marked as inactive while fulfilling an order, which renders it temporarily unable to accept new orders until the current one is completed. As a result, the number of active agents $N_t$ varies across time slots.
    
    \item \textbf{State $\mathcal{S}$}: To support feasible order dispatch decisions, we define the state \( S \) from three dimensions. The state of agent \( k \) at time slot \( t \) is defined as \( s_t^k = \{\mathrm{ST}_t^k, \mathrm{DV}_t^k, \mathrm{CON}_t^k\} \), where \( \mathrm{ST}_t^k \) is the spatial-temporal state, including the current region \( r \), time period \( p \), and coordinates. \( \mathrm{DV}_t^k \) represents the driver state, which encodes the driver's preferences. \( \mathrm{CON}_t^k \) is the context state, capturing global supply-demand conditions, weather, and traffic information.
    
    \item \textbf{Action $\mathcal{A}$}: 
    Agents in our system can perform one of the three action types: accepting an order $a_r$, moving to complete an order $a_m$, or cruising $a_c$ (when the vehicle is unoccupied) based on the driver's preferences. 
    
    \item \textbf{Reward $\mathcal{R}$}: The reward $R^i$ for order $i$ includes both passenger waiting time and fairness, denoted as:
    \begin{equation} \footnotesize \label{eq:4}
        \begin{aligned}
            r^i_t = & - (1 - \alpha) WT(i|u,v) \\
            & - \alpha \left( \frac{1}{K_{u,v}} \sum^{K_{u,v}}_{k=1} (WT(k|u,v) - WT_c(u,v))^2 \right)
        \end{aligned}
    \end{equation}
    \noindent The reward function in our system consists of two components. The first term, \( WT(i \mid u, v) \), reflects system efficiency by representing the waiting time of order \( i \) in region \( u \) during period \( v \). Shorter waiting times yield higher rewards. The second term,  $\frac{1}{K_{u,v}} \sum_{k=1}^{K_{u,v}} (WT(k \mid u, v) - WT_c(u, v))^2$,  serves as a fairness regularization term. Here, \( K_{u,v} \) denotes the total number of passengers in region \( u \) during period \( v \). This term captures the variance between actual waiting times \( WT(k \mid u, v) \) and the dynamic fairness benchmark \( WT_c(u, v) \), encouraging equitable service across different contexts. To balance efficiency and fairness, we introduce a hyperparameter \( \alpha \) to modulate the weight between these two components. In this framework, only the agent fulfilling order \( i \) receives the reward \( r^i_t \).

    \item \textbf{Cost $\mathcal{C}$}: In our setting, each agent $k$ will be assigned a positive region set $\mathcal{H}^{+}_k$, a neutral region set $\mathcal{H}^{0}_k$, and a negative region set $\mathcal{H}^{-}_k$ based on its historical operating locations (i.e., preference).
    Drawing from the negativity effect principle \cite{rozin2001negativity}, negative experiences often exert a stronger influence on our psychological state than equally significant positive or neutral ones. Hence, we impose a cost $c^i_t$ on agent $k$ when an order dispatch leads it to a destination within its negative region set $\mathcal{H}^{-}_k$. The magnitude of this cost $c^i_t$ is determined by the distance between the order's destination and the nearest preferred location within $\mathcal{H}^{+}_k$. 
\end{itemize}

The goal of the defined CMDP $\mathcal{G}$ is to optimize the long-term cumulative reward $J_r(\pi_\theta)$ while ensuring that the cumulative cost $J_c(\pi_\theta)$ remains below a predetermined threshold $\xi$. Given the centralized nature of the order dispatch system overseeing all agents, we adopt a strategy of centralized training with decentralized execution \cite{sharma2021survey} to reduce computational complexity. This strategy emphasizes the cumulative reward and cost across all agents, rather than focusing on the outcomes of individual agents. Therefore, our long-term cumulative reward $J_r(\pi_\theta)$ and cumulative cost $J_c(\pi_\theta)$ can be represented as:
\begin{align}
   & J_r(\pi_\theta) = \mathop{E}_{\substack{\tau \sim \pi_\theta}} \textstyle \sum_{t=0}^{T} \textstyle \sum_{i=1}^{O_t} \left[(\gamma_r)^t r^i_t\right]  \label{eq:7} \\
   & J_c(\pi_\theta) = \mathop{E}_{\substack{\tau \sim \pi_\theta}} \textstyle \sum_{t=0}^{T} \textstyle \sum_{i=1}^{O_t} \left[(\gamma_c)^t c^i_t\right] \label{eq:8} 
\end{align}
where $T$ denotes the total number of time slots within one episode, and $O_t$ represents the total number of orders in time slot $t$. During each slot, we dispatch order $i$ to an agent according to the strategy $\pi_\theta$, and the agent receives a corresponding reward $r^i_t$ and cost $c^i_t$.
\begin{figure*}[htbp] 
    \centering
    \includegraphics[width=0.95\linewidth, keepaspectratio=true]{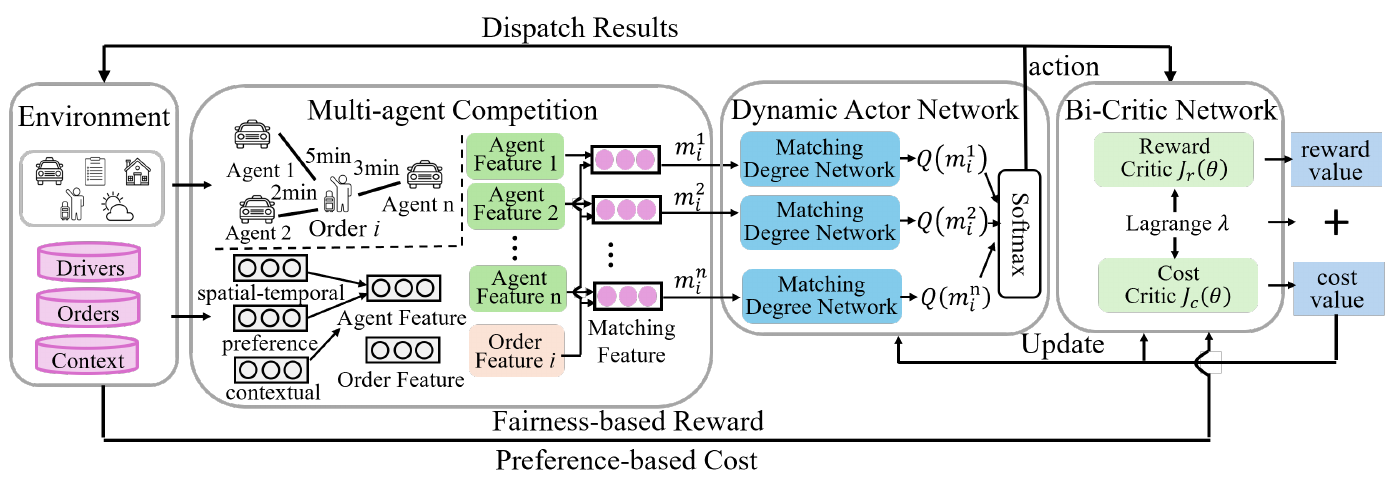}
    \caption{The Overall Framework of the Proposed Habic Method}
    \label{fig:framework}
\end{figure*}

Our objective of maximizing the cumulative reward $J_r(\pi_\theta)$ over an episode and ensuring that the cumulative cost $J_c(\pi_\theta)$ does not exceed a predetermined value $\xi$ can be denoted as Eq.~\ref{eq:9}. It indicates we aim to enhance system efficiency (i.e., reduce the total waiting time of all passengers) and improve passenger fairness without disproportionately compromising driver preferences. 
\begin{equation} \label{eq:9}
\max_{\pi_\theta \in \Pi} J_r(\pi_\theta) \quad s.t. J_c(\pi_\theta) \leq \xi
\end{equation}

\subsection{Optimization Objective Conversion} \label{Conversion}
Directly solving the constrained problem in an MDP is challenging, so we further convert Eq.~\ref{eq:9} into the Lagrangian form:
\begin{equation} \label{eq:10}
    L(\theta,\lambda) = J_r(\theta) - \lambda(J_c(\theta) - \xi)
\end{equation}
\begin{equation} \label{eq:11}
    \max\limits_{\theta}\min\limits_{\lambda} L(\theta,\lambda)
\end{equation}
where the $\lambda \in \mathbb{R}^+$ is the Lagrange multiplier, which is a positive real number. The objective of the above Eq.~\ref{eq:11} aims to find the global optimal saddle point $(\theta^*,\lambda^*)$.
Since $\theta^*$ is the optimal value, the $\theta^*$ should satisfy $L(\theta^*,\lambda^*) \geq L(\theta,\lambda^*)$, $\forall \theta \in \mathbb{R}$. Similarly, $\lambda^*$ should satisfy $L(\theta^*,\lambda^*) \leq L(\theta^*,\lambda)$, $\forall \lambda \in \mathbb{R}^+$. Finally, $\forall (\theta,\lambda)$, we obtain:
\begin{equation} \label{eq:12}
    L(\theta^*,\lambda) \geq L(\theta^*,\lambda^*) \geq L(\theta,\lambda^*)
\end{equation}
However, optimizing the two parameters simultaneously is computationally intractable, especially for $\theta$ that is described by a deep neural network. Therefore, we alternatively optimize the two parameters by fixing one and updating the other until convergence.
We obtain the final $\theta^*$ and $\lambda^*$ when both of them satisfy that:
\begin{equation} \label{eq:13}
   H = \{(\theta^*,\lambda^*)|\ ||\theta^* - \theta^{*-}|| \leq \epsilon_1,||\lambda^* - \lambda^{*-}|| \leq \epsilon_2\}
\end{equation}
Where $\theta^{*-}$ and $\lambda^{*-}$ are the values of the previous values before achieving convergence. In the next part, we will show how we solve this optimization problem with MARL.

\subsection{Harmonization-oriented Actor-Bi-Critic}
In this section, we design a new MARL algorithm called Habic (i.e., \textbf{H}armonization-oriented \textbf{A}ctor-\textbf{Bi}-\textbf{C}ritic) to solve the above-defined problem. There are three key components in the Habic:
(i) A multi-agent competition mechanism, which is designed to provide information for decision-making by generating matching features between orders and drivers. (ii) A dynamic Actor network, which is designed to alternately update the policy parameter $\theta$ and the Lagrange parameter $\lambda$ based on the matching features to make decisions in the multi-agent environment.
(iii) A Bi-Critic network, which is utilized to evaluate the values of accumulated reward $J_r(\pi_\theta)$ and accumulated cost value $J_c(\pi_\theta)$ simultaneously. An overall framework of Habic is shown in Fig.~\ref{fig:framework}.

\subsubsection{Multi-agent Competition Mechanism}
We consider drivers within a certain range of an order to compete for it. The driver selected by the Actor will accept the order, and other drivers will keep their original actions, i.e., staying or cruising. Since the number of drivers around an order is dynamic, the candidate agent set and action space are also dynamic. Each agent in the set generates a matching feature $m_i^k = \{s^k_t, o^i_t\}$ to compete for the order, which includes the state $s^k_t$ of agent $k$ and the state $o^i_t$ of order $i$. The details of $s^k_t = \{{ST}_t^k, {DV}_t^k, {CON}_t^k\}$ can be seen in Sec.~\ref{Sec:3.3}, and $o^i_t = \{OR_i, DE_i, u, v\}$ describes order $i$'s information, including the pickup location $OR_i$ and drop-off location $DE_i$ (represented by longitude and latitude), region $u$, and period $v$. As shown in Fig.~\ref{fig:framework}, $n$ matching features are fed into the Actor network to help make dispatch decisions.

\subsubsection{Dynamic Actor Network for Decision}
In this part, we will first introduce the decision process in Actor, and then show how we update the parameter $\theta$ for the policy function $\pi_\theta(a|s)$ and $\lambda$ for the Lagrange multiplier. We regard the Actor to be equivalent to the policy function $\pi_\theta(a|s)$ to make order dispatch decisions. Assuming there are $n$ candidate agents in the agent set and the matching feature set is $M=\{m_i^1,m_i^2,\cdots,m_i^n\}$, the probability of choosing the $k$th agent can be represented as:
\begin{equation}
    \pi_\theta(a_k|s) = \frac{exp(Q(m_i^k))}{\sum^{n}_{j=1} exp(Q(m_i^j))}
\end{equation}
Where $Q(m_i^k)$ is the matching degree neural network to calculate the matching degree between the order $i$ and the agent $k$. $\pi_\theta(a^k|s)$ means the probability of selecting agent $k$ from the candidate agent set. After calculating the probability of selecting each agent, we can obtain the policy function $\pi_\theta(a|s)$ for the order dispatch decisions.

After executing our order dispatch strategy based on the $\pi_\theta(a|s)$, we can collect the a set of transitions $((s^{1}_{t},\cdots,s^{n_t}_{t}),a_t^k,r^{k}_{t},c^{k}_{t},s^{k}_{t+1})$ and use them as the training data. Based on the gradient search procedure, we can obtain the updating rules for $\theta$ and $\lambda$ as follows:
\begin{align}
     \theta_{n+1} & = \theta_n - \eta_\theta \nabla_{\theta_n}(- L(\theta_n,\lambda_n)) \nonumber \\ 
    & = \theta_n + \eta_\theta [\nabla_{\theta_n} J^R(\pi_\theta) - \lambda_n \nabla_{\theta_n}J^C(\pi_\theta)] \label{eq:15} \\ 
    \lambda_{n+1} & = max\left( 0,\lambda_n + \eta_\lambda  \nabla_{\lambda_n}(- L(\theta_n,\lambda_n))\right) \nonumber \\
    & =  max\left((0,\lambda_n - \eta_\lambda  \nabla_{\lambda_n}(J^C(\pi_\theta)-d))\right) \label{eq:16}
\end{align}
Where $\eta_\theta$ and $\eta_\lambda$ represent the update step sizes for parameters $\theta$ and $\lambda$. As we described in Sec.~\ref{Conversion}, the parameters $\theta$ and $\lambda$ will be updated alternately until they reach the optimal values $\theta^*$ and $\lambda^*$.
However, in Eq.~15 and Eq.~16, the values of $J^R(\pi_\theta)$ and $J^C(\pi_\theta)$ still remain unknown. Therefore, we further design the Bi-Critic network to evaluate their values.

\subsubsection{Bi-Critic Network for Evaluation}
This section introduces how we evaluate the value of $J^R(\pi_\theta)$ and $J^C(\pi_\theta)$. As shown in Eq.~\ref{eq:7} and Eq.~\ref{eq:8}, the $J^R(\pi_\theta)$ and $J^C(\pi_\theta)$ share the same structure, so we will only show the evaluation process for $J^R(\pi_\theta)$, and the evaluation process of $J^C(\pi_\theta)$ is the same. According to \cite{schulman2017proximal}, $J^R(\pi_\theta)$ can be rewritten as:
\begin{equation} \label{eq:17}
   J^R(\pi_\theta) = E_{s \sim D^\pi(s)}E_{a \sim \pi_{\theta^{-}}} \left[\frac{\pi_\theta(a|s)}{\pi_{\theta^{-}}(a|s)}A^R_{\pi_\theta}(s,a)\right]
\end{equation}
Where $D^\pi(s)$ is the state visitation distribution, which can be described as the average probability of the state $s$ appearing at each moment in the trajectory. $\pi_{\theta^{-}}$ is the old strategy in the last cycle and $\pi_\theta$ is the new strategy waiting to be updated in this cycle, which means $\pi_{\theta^{-}}$ and $\pi_\theta$ are equivalent to $\pi_{\theta_{n}}$ and $\pi_{\theta_{n+1}}$ in Eq.~\ref{eq:15}. $A^R_{\pi_\theta}(s,a)$ is the Advantage function, which can be considered as another version of Q-value with lower variance by taking the state-value off as the baseline. In Habic, we calculate $A^R_{\pi_\theta}(s,a)$ by utilizing the Generalized Advantage Estimation (GAE) method \cite{schulman2015high}, which can be described as:
\begin{equation} \label{eq:18} 
    A^R_{\pi_\theta}(s_t,a_t) = \sum^{\infty}_{l=0}(\gamma_r\psi)^l \left(r_t + \gamma_r V^R_{\pi_\theta}(s_t) - \gamma_r V^R_{\pi_\theta}(s_{t+1})\right)
\end{equation}
Where $\psi \in [0,1]$ is a hyper-parameter in GAE, and $V^R_{\pi_\theta}(s_t)$ is the value function to describe the value of state $s_t$ when following a policy $\pi_\theta$.

In Eq.~\ref{eq:17}, we adopt the off-policy strategy \cite{brandfonbrener2021offline} by using the old strategy $\pi_{\theta^{-}}$ to collect the data and update the new strategy parameter $\theta$, so the difference between the new strategy and the old strategy will not be too large. We also leverage the KL-divergence \cite{kullback1951kullback} to restrict the updating range of $\theta$, which is shown in Eq.~\ref{eq:19}:
\begin{equation} \label{eq:19}
    E_{s \sim D^\pi(s)}\left[D_{KL}(\pi_{\theta^{-}}(\cdot|s),\pi_{\theta}(\cdot|s)) \right] \leq \delta
\end{equation}
By combining Eq.~\ref{eq:17} and Eq.~\ref{eq:19}, we can obtain the final describe of $J^R(\pi_\theta)$ based on the PPO-Clip \cite{jayant2022model}:
\begin{equation} \label{eq:20}
\begin{split}
J^R(\pi_\theta) & = E_t \bigg[ \min\bigg(\frac{\pi_\theta(a_t|s_t)}{\pi_{\theta^{-}}(a_t|s_t)}A^R_{\pi_\theta}(s_t,a_t),\\
& clip(\frac{\pi_\theta(a_t|s_t)}{\pi_{\theta^{-}}(a_t|s_t)},1-\epsilon,1+\epsilon)A^R_{\pi_\theta}(s_t,a_t) \bigg) \bigg]
\end{split}
\end{equation}
Where $clip(x,a,b) = max(min(x,b),a)$, which means restricting $x$ within the range $[a, b]$. Using the Eq.~\ref{eq:20}, we can estimate the value of $J^R(\pi_\theta)$ based on the Advantage function $A^R_{\pi_\theta}(s_t,a_t)$, which can be represented by $V^R_{\pi_\theta}(s_t)$ based on Eq.~\ref{eq:18}. Therefore, we build the Reward Critic network to calculate $V^R_{\pi_\theta}(s_t)$ with the parameter $\psi_r$. We denote the Reward Critic network as $V^R_{\psi_r}(s_t)$. Similarly, the Cost Critic network can be denoted as $V^C_{\psi_c}(s_t)$ with the parameter 
$\psi_c$. The update rules for parameter $\psi_r$ and $\psi_c$ are:
\begin{align}
    {\psi_r} \leftarrow  {\psi_r} - \eta_{{\psi_r}}\nabla\textstyle\sum
[r_t + \gamma_rV^R_{{\psi_r}}(s_{t+1}) - V^R_{{\psi_r}}(s_{t})] \label{eq:21}\\
    {\psi_c} \leftarrow  {\psi_c} - \eta_{{\psi_c}}\nabla\textstyle\sum
[r_t + \gamma_cV^C_{{\psi_c}}(s_{t+1}) - V^C_{{\psi_c}}(s_{t})] \label{eq:22}
\end{align}

To summarize, we build a Reward Critic network $V^R_{\psi_r}(s_t)$ and a Cost Critic network $V^C_{\psi_c}(s_t)$ to estimate $V^R_{\pi_\theta}(s_t)$ and $V^C_{\pi_\theta}(s_t)$, respectively. By updating $\psi_r$ and $\psi_c$ using Eq.~\ref{eq:21} and Eq.~\ref{eq:22}, we obtain the accumulated estimates. The values of $V^R_{\pi_\theta}(s_t)$ and $V^C_{\pi_\theta}(s_t)$ are then used to compute the accumulated reward $J^R(\pi_\theta)$ and cost $J^C(\pi_\theta)$ based on Eq.~\ref{eq:18} and Eq.~\ref{eq:20}. After obtaining these estimates, we use them to update the parameters $\theta$ and $\lambda$, where the converged $\theta^*$ serves as the optimal solution to our optimization objective in Eq.~\ref{eq:9}, improving passenger fairness without disproportionately compromising driver preferences.

\section{Evaluation} \label{Sec6-EVALUATION}

\begin{table*}[t] \label{Tb:performance}
    \small
    \centering
    \begin{threeparttable}
        \renewcommand{\arraystretch}{1.2}
        \setlength\tabcolsep{1.9mm}{
        \begin{tabular}{c|c|cc|c|c|cc|c}
            \hline
            Cities & \multicolumn{4}{c|}{Shenzhen} & \multicolumn{4}{c}{NYC} 
            \\
            \hline
            \multirow{2}{*}{Methods} & \multicolumn{1}{c|}{Efficiency} & \multicolumn{2}{c|}{Fairness} & \multicolumn{1}{c|}{Preference} & \multicolumn{1}{c|}{Efficiency} & \multicolumn{2}{c|}{Fairness} & \multicolumn{1}{c}{Preference} \\ 
            \cline{2-9} & $DAPWT$ & $DPF_{inter}$ & $DPF_{intra}$ & $DPVR$ & $DAPWT$ & $DPF_{inter}$ & $DPF_{intra}$ & $DPVR$   \\
            \hline \hline
             DQND  & 2.51$\pm$0.34 & 5.26$\pm$1.75 & 6.05$\pm$1.57 & 3.33$\pm$3.56 & 3.43$\pm$0.87 & 7.81$\pm$2.23 & 9.92$\pm$2.41 & 4.25$\pm$3.29  \\   AC-bgm  & 4.03$\pm$0.85 & 6.92$\pm$1.17 & 7.73$\pm$1.34 & -2.25$\pm$3.31 &  4.77$\pm$1.52 & 10.21$\pm$2.37 & 9.54$\pm$2.11 & 2.53$\pm$3.11 \\  IPPO  & -0.99$\pm$0.78 & 2.34$\pm$1.03 & 2.55$\pm$1.49 & 3.51$\pm$3.25 & -3.01$\pm$1.94 & 4.41$\pm$3.05 & 5.87$\pm$2.08 & -0.68$\pm$2.34\\  MAPPO  & 1.11$\pm$0.94 & 3.47$\pm$1.94 & 4.12$\pm$0.98 & 3.11$\pm$3.37 & 2.26$\pm$1.99 & 8.28$\pm$2.03 & 7.92$\pm$2.56 & -1.15$\pm$3.48 \\   CPO & 4.51$\pm$1.12 & 7.73$\pm$1.26 & 8.52$\pm$1.94  & 7.75$\pm$2.95 & 4.72$\pm$1.03 & 9.88$\pm$1.88 & 12.73$\pm$2.47 & 12.19$\pm$5.38  \\  Lag-TRPO  & 3.78$\pm$1.32 & 8.55$\pm$1.46 & 9.77$\pm$1.54 & 12.74$\pm$3.29 & 5.24$\pm$2.17 & 10.38$\pm$2.69 & 15.42$\pm$3.02 & 15.81$\pm$4.42 \\  HCRide-AF & 5.31$\pm$1.45 & 7.00$\pm$1.82 & 7.25$\pm$1.69 & 19.23$\pm$3.68 & 7.01$\pm$1.98 & 10.01$\pm$2.77 & 10.86$\pm$2.98 & 22.45$\pm$4.01 \\   HCRide-MMF & 5.42$\pm$1.17 & 9.98$\pm$1.68 & 11.36$\pm$1.92 & \textbf{19.79$\pm$3.03} & 7.12$\pm$2.08 & 13.25$\pm$3.01 & 14.34$\pm$3.09 & 23.85$\pm$4.21  \\ HCRide & \textbf{5.48$\pm$1.13} & \textbf{13.39$\pm$1.71}  & \textbf{16.67$\pm$1.46} & 19.52$\pm$3.87 & \textbf{7.15$\pm$2.03} & \textbf{15.11$\pm$2.88} & \textbf{19.98$\pm$3.03} & \textbf{24.41$\pm$3.92} \\ \hline \hline
        \end{tabular}}
    \end{threeparttable}
\caption{
Comparison of different methods across multiple metrics. The symbol \% is omitted in the table. 
\textit{DAPWT}, \textit{DPF}\textsubscript{\textit{inter}}, \textit{DPF}\textsubscript{\textit{intra}}, and \textit{DPVR} represent the decreased ratios in Average Passenger Waiting Time (\textit{APWT}), inter-region Passenger Fairness (\textit{PF}\textsubscript{\textit{inter}}), intra-region Passenger Fairness (\textit{PF}\textsubscript{\textit{intra}}), and Preference Violation Rate (\textit{PVR}), respectively.
Lower values indicate better performance. All values are computed relative to the benchmark baseline \textit{MD}~\protect\cite{zhang2017taxi}. 
The best performance is highlighted in bold.
}
    \label{tb:main performance}
\end{table*}

\subsection{Evaluation Methodology}

\noindent \textbf{Data:}
We evaluate our HCRide on two real-world ride-hailing datasets from the Chinese City Shenzhen and New York City (NYC). The Shenzhen dataset includes 1.07 million orders served by 1,200 ride-hailing vehicles from 03/2021 to 06/2021. The NYC dataset includes 214k orders served by 800 ride-hailing vehicles from 01/2024 to 02/2024. 

\noindent \textbf{Baselines:}
We compare our HCRide with five different categories of baselines: (1) Myopic dispatching method: $MD$ \cite{zhang2017taxi}. This method aims to minimize the total waiting time for all the passengers in one slot without future consideration. The method $MD$ will be considered the benchmark to be compared with all other methods in Table~\ref{tb:main performance}. (2) Single-agent RL methods: DQND \cite{mnih2013playing}, AC-bgm \cite{wang2023time}. (3) Multi-agent RL methods: IPPO \cite{de2020independent}, MAPPO \cite{yu2022surprising}. (4) Constrained RL methods: CPO \cite{achiam2017constrained}, Lag-TRPO \cite{ray2019benchmarking}. Compared to the previous two types of methods, the constrained RL methods introduce the cost and constraint. (5) Variants of our HCRide considering different fairness definitions: HCRide-AF with absolute fairness \cite{suhr2019two}, HCRide-MMF with max-min fairness \cite{sun2022optimizing}.

\noindent \textbf{Metrics:}
We define three categories of metrics to evaluate the performance of system efficiency (Average Passenger Waiting Time $APWT$), passenger fairness (inter-region fairness $PF_{inter}$ and intra-region fairness $PF_{intra}$ based on the variance of passenger waiting time), and driver preference (Preference Violation Rate $PVR$), which evaluates the proportion of orders assigned to non-preferred regions of drivers, i.e., the negative region set $H^-_k$.

\subsection{Overall Performance} \label{Overall Performance}
As shown in Table \ref{tb:main performance}, we compare our HCRide with all eight baselines on the two datasets. 

\subsubsection{System Efficiency} \label{efficiency}
We evaluate the system efficiency using the Average Passenger Waiting Time $APWT$. As shown in Table \ref{tb:main performance}, our HCRide outperforms all other baselines. Using the Shenzhen dataset as an example, our HCRide reduces the average passenger waiting time by 5.48\% compared to the benchmark baseline $MD$ and outperforms the state-of-the-art method Lag-TRPO by 1.77\% $=(5.48\% - 3.78\%)/(100\% - 3.78\%)$ in the whole day and 1.53\% in the morning rush hour. Compared to Lag-TRPO, HCRide utilizes a more efficient updating method PPO-clip \cite{jayant2022model}, thereby achieving better convergence. HCRide-AF and HCRide-MMF are variants of HCRide with different fairness settings. Since there are no changes to the efficiency settings, their efficiency performances are similar. In particular, single-agent methods DQND and AC-bgm can achieve better performance than multi-agent methods IPPO and MAPPO. One possible reason is that we use the discrete extension for the single-agent RL methods, which incorporates our multi-agent competition mechanism. This allows these methods to learn the value of order-passenger pair from every discrete order dispatch behavior, providing abundant training transactions to guide learning the matching degree between orders and passengers. However, multi-agent RL methods like IPPO and MAPPO can learn from the operation trajectories of each driver. When there are a large number of agents (e.g., over 1,000), it will bring a high variance for training. Additionally, it is challenging for the Actor to learn the competition among different drivers since they are trained independently.

\subsubsection{Passenger Fairness} \label{fairness}
\begin{figure}[t]\centering
    \centering    \includegraphics[width=0.88\linewidth, keepaspectratio=true]{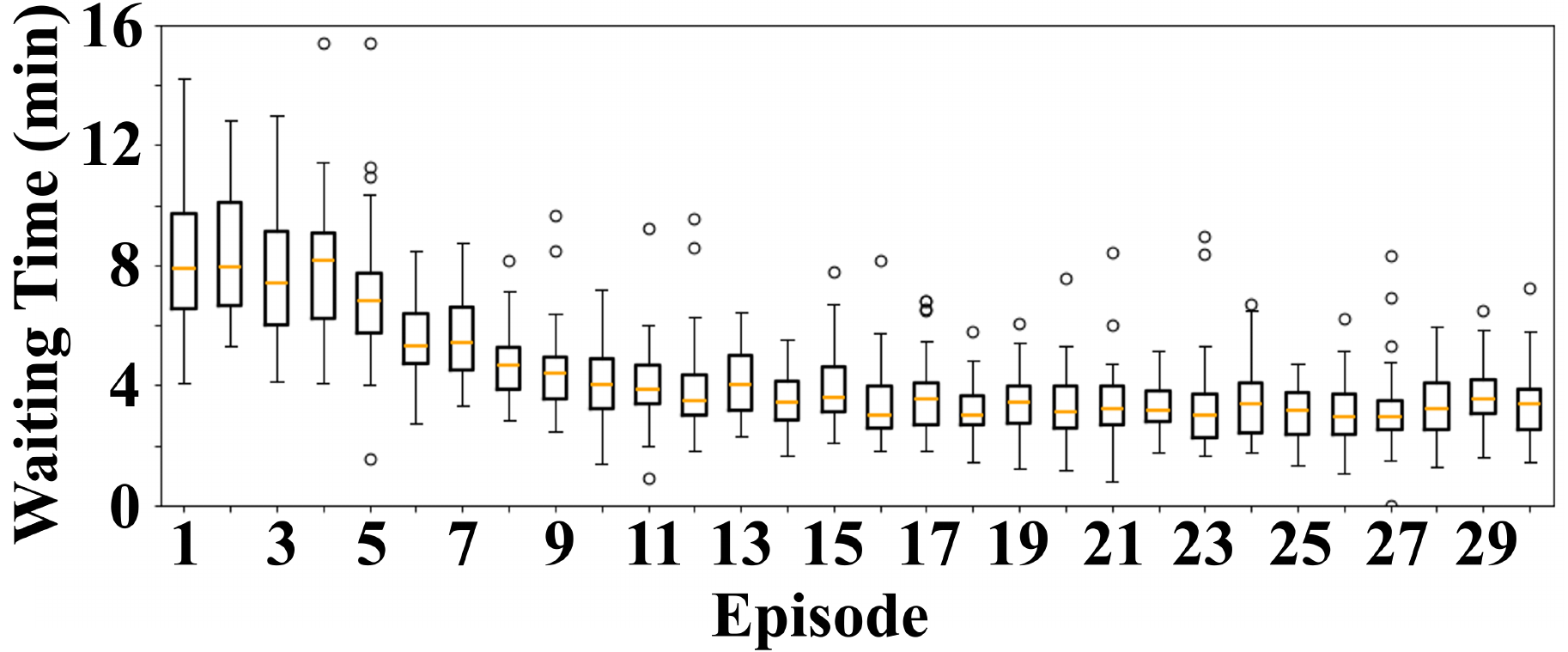}
    \caption{Average Waiting Time Distribution for All Regions}
    \label{fig:inter-region fairness}
\end{figure}

We evaluate passenger fairness from both inter-region and intra-region levels with metrics $DPF_{inter}$ and $DPF_{intra}$. As shown in Table \ref{tb:main performance}, our HCRide notably improves both metrics. For single-agent RL algorithms such as AC-bgm and constrained RL algorithms like Lag-TRPO, although they utilize the same fairness-based reward function as HCRide, they are less effective due to poor exploration and update ability. We also compare the performance of our spatio-temporal-aware fairness definition with two other widely used fairness definitions: absolute fairness \cite{zhou2002relationship} and max-min fairness \cite{sun2022optimizing}. The results show that our HCRide can achieve better fairness performance compared to HCRide-AF and HCRide-MMF. A possible reason is that our spatio-temporal-aware fairness definition focuses on more fine-grained local information across different spatio-temporal contexts. We also provide the visualization results for inter-region fairness in Fig.~\ref{fig:inter-region fairness}, which shows the distribution of the average passenger waiting times of all regions on each day during the training process. We find that the variance of average passenger waiting time between different regions decreases during the training process, and the system eventually converges to be close to the fairness benchmark.

\subsubsection{Driver Preference} \label{preference}

For driver preference, we focus on the percentage of dispatched orders that differ from driver preferences. Using the Shenzhen dataset as an example, the driver preference violation rate reaches 18.21\% for the benchmark $MD$. In this experiment, we set a predetermined violation rate of 15\% and expect it to decrease during the training process, eventually converging below this value. The predetermined violation rate can also be set to other values based on operators' goals. From Table \ref{tb:main performance}, we find that HCRide achieves the best performance on the two datasets. In contrast, baselines such as DQND, AC-bgm, IPPO, and MAPPO, which fail to consider driver performance in order dispatch decisions, show performance similar to the myopic dispatch strategy. Constrained RL baselines such as CPO and Lag-TRPO outperform other non-constrained RL baselines but still fall short compared to our HCRide. A key reason is that our HCRide has a multi-agent competition mechanism to improve the sampling efficiency and a Bi-Critic to provide direct reward value and cost value estimation for the matching degree.

\section{Related Work} \label{Sec2-RELATED WORK}
We divide order dispatch work into efficiency-oriented order dispatch and human-centered order dispatch.

\textbf{Efficiency-oriented Order Dispatch:} Most existing studies prioritize efficiency without considering human factors such as fairness and preference. Xu et al. \cite{xu2018large} use the DRL to solve sequential dispatch problems by building the global Q-function for orders and passengers. Sadeghi et al. 
\cite{sadeghi2022reinforcement} propose a scalable RL dispatching algorithm and conduct both offline evaluation and online evaluation. Recently, some works have begun to pay attention to the transferability of algorithm efficiency across various platforms. Wang et al. \cite{wang2018deep} use the transfer learning method to make DRL-based order dispatch algorithms more adaptive in different cities. Wang et al. \cite{wang2022fed} propose a federated learning algorithm to improve the reliability of dispatching data during cross-platform processes.

\textbf{Human-centered Order Dispatch:} In recent years, human-centered design has attracted much interest, and more and more works focus on fairness. Sühr et al. \cite{suhr2019two} propose an order dispatch method considering two-sided fairness for both driver and passenger. Lu et al. \cite{lu2021efficiency} introduce the queueing theory to solve the long waiting time problem for passengers and make a trade-off between efficiency and fairness. There are also some other works focusing on human preference. Carvalho et al. \cite{de2022satisfying} propose a multi-agent multi-objective optimization approach to satisfy user preferences in ridesharing services. Li et al. \cite{li2021preference} introduce the mutual information-based approach to solve the preference-aware group task assignment in spatial crowdsourcing. 

To our knowledge, our HCRide is the first order dispatch system that harmonizes both passenger fairness and driver preference for human-centered ride-hailing services. 

\section{Conclusion}

Motivated by insights from our previous qualitative study and data-driven analysis, in this paper, we design a human-centered ride-hailing order dispatch system called HCRide, which aims to improve both system efficiency and passenger fairness in terms of waiting time without compromising driver preferences. In HCRide, spatio-temporal-aware fairness and preference are formally defined, and we design a novel multi-agent reinforcement learning algorithm called harmonization-oriented Actor-Bi-Critic, which includes a multi-agent competition mechanism, a dynamic Actor network, and a Bi-Critic network to optimize system efficiency and passenger fairness with driver preferences as constraints. Extensive evaluations on two datasets show our HCRide effectively improves system efficiency by 2.02\%, inter-region fairness by 5.28\%, intra-region fairness by 5.39\%, and driver preference by 10.21\% compared to baselines.

\section*{Acknowledgments}
We sincerely thank all anonymous reviewers for their insightful comments and valuable suggestions. This work is partially supported by Florida State University, National Science Foundation under Grant Numbers 2411152, 2427915, and 2318697.

\bibliographystyle{named}
\bibliography{ijcai25}

\end{document}